\documentclass[11pt]{article}
\usepackage{graphicx}
\usepackage[boxruled,vlined,linesnumbered]{algorithm2e}
\usepackage{amsbsy}
\usepackage{amssymb}
\begin{document}

\title{An Improved Discrete Bat Algorithm for Symmetric and Asymmetric Traveling Salesman Problems}

\author{Eneko Osaba$^1$, Xin-She Yang$^2$, Fernando Diaz$^1$, Pedro Lopez-Garcia$^1$, Roberto Carballedo$^1$ \\[10pt]
1) Deusto Institute of Technology (DeustoTech),  \\
University of Deusto, Av. Universidades 24, Bilbao 48007, Spain \\
2) School of Science and Technology, \\
Middlesex University, Hendon Campus, London, NW4 4BT, United Kingdom \\
}

\date{\hrule
{\bf Citation Detail:} \\
E. Obsab, X. S. yang, F. Diaz, P. Lopez-Garcia, R. Carballedo,
An ,Improved Discrete Bat Algorithm for Symmetric and Asymmetric Traveling Salesman Problems, {\it Engineering Applications of Artificial Intelligence},
48 (1), 59-71 (2016).
\hrule
 }

\maketitle

\begin{abstract}
Bat algorithm is a population metaheuristic proposed in 2010 which is based on the echolocation or bio-sonar characteristics of microbats. Since its first implementation, the bat algorithm has been used in a wide range of fields. In this paper, we present a discrete version of the bat algorithm to solve the well-known symmetric and asymmetric traveling salesman problems. In addition, we propose an improvement in the basic structure of the classic bat algorithm. To prove that our proposal is a promising approximation method, we have compared its performance in 37 instances with the results obtained by five different techniques: evolutionary simulated annealing, genetic algorithm, an island based distributed genetic algorithm, a discrete firefly algorithm and an imperialist competitive algorithm. In order to obtain fair and rigorous comparisons, we have conducted three different statistical tests along the paper: the Student's $t$-test, the Holm's test, and the Friedman test. We have also compared the convergence behaviour shown by our proposal with the ones shown by the evolutionary simulated annealing, and the discrete firefly algorithm. The experimentation carried out in this study has shown that the presented improved bat algorithm outperforms significantly all the other alternatives in most of the cases.
\end{abstract}

\section{Introduction}
\label{sec:intro}

Combinatorial optimization is one of the most studied fields in artificial intelligence, optimization, logistics, and other applications. Multiple research works are published annually in this area, both in journals (\cite{kasperski2015combinatorial}), and conferences (\cite{bezerra2014automatic}), and also in books (\cite{levin2015system}). Different sort of problems exist within this kind of optimization, including the routing problems as one of the most appealed. It is noteworthy that the most used and well-known routing problems are the Traveling Salesman Problem (TSP) (\cite{lawler1985traveling}), and the Vehicle Routing Problem (VRP) (\cite{christofides1976vehicle}), which are the focus of a huge amount of studies in the literature (\cite{groba2015solving,bortfeldt2015hybrid}).

The main reasons for the popularity and importance of the routing problems are two folds: the social interest they generate, and their inherent scientific interest. On the one hand, routing problems are normally designed to deal with real world situations related to the transport or logistics. This is the reason why their efficient resolution entails a profit, either social or business one. On the other hand, most of the problems arising in this field have a great computational complexity. Being NP-Hard, the resolution of these problems is a major challenge for the scientific community.

In line with this, diverse appropriate approaches can be found in the literature to address this kind of problems efficiently. Arguably the most successful techniques are the exact methods (\cite{laporte1992traveling, laporte1992vehicle}), heuristics and metaheuristics. In this paper, we focus our attention in the last ones. Some classical examples of metaheuristics can be the simulated annealing (SA) (\cite{simulatedAnnealing}), and the tabu search (\cite{tabuSearch}), as local search-based methods, and genetic algorithm (GA) (\cite{genetic1,genetic2}), particle swarm optimization \\
(\cite{PSO}, \cite{PSO1}), and ant colony optimization (\cite{ACO}) as population-based ones. Despite having been proposed many years ago, these techniques remain successful in scientific community nowadays, being the cornerstone of multiple studies  \\ (\cite{rodriguez2015rwa}, \\ \cite{cao2015tabu}, \cite{inkaya2015ant}).

In spite of the existence of these classic approaches, the design and implementation of novel meta-heuristics for addressing optimization and routing problems is a hot topic for the scientific community today. For this reason, many different metaheuristics have been proposed in the last decade, which have been successfully applied to various problems and fields. Some examples of these techniques are the artificial bee colony, proposed in 2005 by Karaboga (\cite{ABC,ABC1,ABC2},\\  \cite{ABC3}), the imperialist competitive algorithm, presented by Gargari and Lucas in 2007 (\cite{imperialist}), or the firefly algorithm, proposed by Yang in 2009 (\cite{yang2009firefly}).

For this reason, this paper is focused on one metaheuristic proposed few years ago, called Bat Algorithm (BA). This population technique was proposed by Yang in 2010 (\cite{yang2010new}), and it is based on the echolocation behavior of microbats, which can find their prey and discriminate different kinds of insects even in complete darkness. As can be read in several surveys (\cite{yang2013bat,parpinelli2011new}), since its proposal the BA has been successfully applied to different optimization fields and problems. Additionally, the fact that many research works focused on BA are being currently published confirms that BA still attracts a lot of interest (\cite{fister2015planning,meng2015novel}). As we have mentioned, the BA has been applied to many different optimization problems, anyway, it has been rarely applied to any routing problem (\cite{BATSP}). This lack of works and the growing interest in the BA by the scientific community has motivated this work.

In this work, we present a discrete BA for solving routing problems. Being one of the first times that BA addresses this sort of problems, two of the most studied routing problems have been used for the experimentation: the TSP and the Asymmetric TSP (ATSP). Besides this, we also present an improved version of the basic BA (IBA), which outperforms the basic versions significantly.

The main objective of this study is to prove that the IBA is a promising approximation method for the TSP and ATSP. To prove that, we compare the results obtained by the IBA with the ones obtained by two basic versions of the BA, and with the ones obtained by five different well-known metaheuristics: GA, evolutionary simulated annealing (ESA) (\cite{ESA}), the Island based Distributed Genetic Algorithm (IDGA) (\cite{IDGA}), a Discrete Firefly Algorithm (DFA), and a Discrete Imperialist Competitive Algorithm (DICA). To perform this comparison, 37 different TSP-ATSP instances have been used in the experimentation carried out. Furthermore, with the objective of drawing rigorous and fair conclusions, in addition to the conventional comparison based on the typical descriptive statistics parameters (results average, standard deviation, best result, etc.), we also perform three statistical tests along the paper: the Student's $t$-test, the Holm's test, and the Friedman test.

The remainder of this paper is structured as follows. In the following section (Section \ref{sec:related}), a brief literature related to the BA is presented. In Section \ref{sec:Bat} the basic aspects of the BA are detailed. In Section \ref{sec:TSP} a brief description of the TSP and ATSP is made. Then, our proposed discrete BA and IBA are described in Section \ref{sec:OurBat}. Additionally, the experimentation carried out is described in Section \ref{sec:exp}. Finally, conclusions and future work are explained in Section \ref{sec:conc}.

\section{Related Work}
\label{sec:related}

As we have mentioned in the previous section, the BA is a population algorithm proposed in 2010 by Yang. The basic BA is based on the echolocation or bio-sonar characteristics of microbats, and its first version was proposed for solving continuous optimization problems. Since this first implementation, the BA has been applied in a wide range of fields. Some of these fields are the continuous optimization, in which some additional works have been published apart from to the original one, (\cite{bora2012bat,yang2012bat}), combinatorial optimization (\cite{marichelvam2013solving}), image processing (\cite{zhang2012image}) and clustering problems (\cite{komarasamy2012optimized}).

Besides this, many variations of the basic BA have been proposed in the literature. One example is the Fuzzy Logic BA (FLBA), presented by Khan et al. in 2011 (\cite{khan2011fuzzy}), which introduces some fuzzy logic mechanisms in the basic structure of the BA. This first FLBA was proposed as method for ergonomic screening of office workplaces. Another example of FLBA can be seen in (\cite{perez2015new}). In this work the authors present a FLBA for dynamical parameter adaption. Other example of BA variation is the chaotic BA (CBA). The first CBA, which uses L\'evy flights and chaotic maps, was proposed by Lin et al. for parameter estimation in dynamic biological systems (\cite{lin2012chaotic}). Furthermore, in 2014 an improved CBA was presented by Abdel-Raouf et al. for solving integer programming problems (\cite{abdel2014improved}). In the same year, Gandomi and Yang proposed a CBA for robust global optimization (\cite{gandomi2014chaotic}). Two other examples of BA variants are the BA with mutation (\cite{zhang2012image}) or the multi-objective BA (\cite{yang2011bat}).

Additionally, some hybrid techniques have been developed using the BA as one of the hybridized methods. In (\cite{pan2015hybrid}), for example, a hybrid particle swarm optimization with BA was developed. This approach, presented by Pan et al. in 2015, was implemented to deal with numerical optimization problems. One year earlier, Nguyen et al. proposed a hybrid bat algorithm with artificial bee colony also for solving the same kind of problems (\cite{nguyen2014hybrid}). On the other hand, Meng et al. presented in 2015 a hybrid BA with differential evolution strategy for addressing constrained optimization problems (\cite{meng2015novel}). Furthermore, Ramawan et al. developed in 2014 a BA hybridized with an artificial neural network. This novel approach was implemented to predict the output power of grid-connected photovoltaic system. At last, in (\cite{roeva2013hybrid}) a method which combines the BA with sequential quadratic programming for facing the parameter identification of an E. coli fed-batch cultivation process model was presented.

In the present work, we develop a discrete version of the BA. Although its first version was designed for continuous problems, the BA has been modified many times in the literature with the intention of addressing discrete optimization problems. In (\cite{nakamura2012bba}), for instance, we can find the first Binary Bat Algorithm (BBA) applied to feature selection problems. Another successful BBA was developed by Mirjalili et al. in 2014 for solving discrete optimization problems (\cite{mirjalili2014binary}). A recently paper published by Fister et al. presents another discrete version of the BA for the correct planning of the sports training sessions (\cite{fister2015planning}). Finally, in (\cite{luo2014discrete}) a discrete BA was developed by Luo et al. for addressing the optimal permutation flow shop scheduling problem, and in (\cite{marichelvam2013solving}) another discrete version of the BA was proposed for solving hybrid flow shop scheduling problems.

In spite of this great amount of research studies, as we have mentioned in the introduction of this paper, the BA has been rarely applied to any routing problem. This lack of studies has been the main motivation that has driven the realization of this work. Anyway, the main novelty of the presented IBA is not only its application field. The technique developed in this work presents the originality of using the Hamming Distance function to measure the distance between two bats of the swarm. This approach has been used previously in other techniques applied to the TSP, proving its good performance (\cite{zhou2014multi}), but it has been never used for any BA. In addition, according to the basic philosophy of BA, all the bats of the swarm perform their movements always in the same way. This strategy is not used in the proposed IBA, where the bats are endowed with certain ``intelligence". In this way, bats employ different movement schemes depending on the point of the solution space in which they are located. This is the first time that this approach is used in the literature.

On the other hand, as can be read in several studies of the literature, since its formulation, the TSP is one of the standard test problems used in performance analysis of discrete optimization algorithms (\cite{mahi2015new}). Many classic techniques have been applied to the TSP in the last few decades, as genetic algorithms (\cite{grefenstette1985genetic,larranaga1999genetic}), simulated annealing (\cite{malek1989serial}, \\ \cite{aarts1988quantitative}),
or the tabu search (\cite{fiechter1994parallel,knox1994tabu},  \cite{gendreau1998tabu}). Despite being classical techniques, these methods have been also applied to the TSP in many recent studies (\cite{misevivcius2015using,wang2014hybrid,nagata2012new}). More recent but still well-known techniques have been also widely applied for the TSP, as the ant colony optimization (\cite{dorigo1997ant,jun2012application}), the variable neighborhood search (\cite{carrabs2007variable,burke2001effective}), \\ or the particle swarm optimization (\cite{clerc2004discrete,shi2007particle}). As has been mentioned with respect to the previous ones, these techniques are also being applied nowadays to the TSP (\cite{yao2014genetic,pang2015improved,ariyasingha2015performance}).

Additionally, the TSP has also been used as benchmarking problem for bio-inspired techniques proposed in the last decade. Some of these recent developed meta-heuristics are the firefly algorithm (\cite{jati2015firefly,li2015firefly}), which is based on the social behavior of fireflies and the phenomenon of bioluminescent communication, the Cuckoo Search (\cite{ouaarab2014discrete}), which is inspired by the breeding behaviour of cuckoos, or the Imperialist Competitive Algorithm  \\  (\cite{ardalan2015novel,yousefikhoshbakht2013new}), which is a socio-politically motivated global search strategy, based on the imperialist competition of the countries. Some other examples of these recent developed bio-inspired meta-heuristics which have been applied to the TSP are the artificial bee colony algorithm (\cite{karaboga2011combinatorial}), inspired by the intelligent behaviour of honey bee swarm, or the Honey Bees Mating Optimization (\cite{marinakis2011honey}).

It is also worth mentioning the hybrid techniques developed and tested with the TSP, as the one presented in (\cite{saenphon2014combining}), which combines the well-known ant colony optimization with the gradient search. On the other hand, in (\cite{mahi2015new}) a hybrid method based on a particle swarm optimization, ant colony optimization and 3-opt algorithm is presented and applied to the TSP. Continuing with this concept, in (\cite{chen2011solving}) a technique hybridizing a genetic simulated annealing, an ant colony system and a particle swarm optimization was implemented and used to solve the TSP. Finally, another interesting study is the presented in (\cite{zhao2015simulated}), in which a simulated annealing hybridized with local searches is introduced.

In this study, in order to prove that the proposed meta-heuristic is a promising one, its performance is compared with three classical techniques, GA, ESA and IDGA, and two recently proposed ones, DICA and DFA.

Finally, it is noteworthy that this small set of works listed in this section is only a small sample of all the related work. Due to the high number of related papers, to summarize all the relevant work may be a huge task. For this reason, if any reader wants more information regarding the possible applications of BA, we recommend the reading of the literature review paper presented in (\cite{yang2013bat}). On the other hand, for further information of the TSP and its solving techniques, the work presented in (\cite{anbuudayasankar2014survey}) is recommended.

\section{Bat Algorithm}
\label{sec:Bat}

As we have briefly mentioned in previous sections, the BA is a bio-inspired metaheuristic based on the echolocation system of bats. In nature, bats emit ultrasonic pulses to the surrounding environment with hunting and navigation purposes. After the emission of these pulses, bats listen to the echoes, and based on them they can locate themselves and also locate and identify obstacles and preys. Furthermore, each bat of the swarm is able to find the most ``nutritious" areas performing an individual search, or moving towards a ``nutritious" location previously found by the swarm.

The main idea of the BA is to imitate this echolocation system of the bats. Anyway, some idealized rules have to be taken into account in order to make a proper adaptation (\cite{yang2010new}):

\begin{itemize}
	\item All bats use echolocation to detect the distance, and they have one ``magic ability" that allow them to difference between a prey and an obstacle.
	\item All bats fly randomly with a velocity $v_i$ at position $x_i$ with a fixed frequency $f_{min}$, varying wavelength $\lambda$ and loudness $A_i$ to search a prey. In this idealized rule, we assume that every bat can adjust in an automatic way the frequency (or wavelength) of the emitted pulses, and the rate of these pulses emission $r\in[0,1]$ . This automatic adjustment depends on the proximity of the targeted prey.
	\item In real situations, the loudness of bats emissions can vary in many ways. Nonetheless, we assume that this loudness can vary from a large positive $A_0$ to a minimum constant value $A_{min}$.
\end{itemize}

In Algorithm \ref{alg:BA} the pseudocode of the basic BA is shown. Taking a look to this algorithm we can see that lines 1-6 correspond to the initialization process. First, the objective function has to be defined, and the initial population has to be initialized. We assume that every bat of the population represents one possible solution to the addressed problem. Then, all the parameters related to each bat are initialized and defined. These parameters are the velocity $v_i$, frequency $f_i$, pulse rate $r_i$ and loudness $A_i$.

\begin{algorithm}[tb]
	 \SetAlgoLined
		Define the objective function $f(x)$\;
		Initialize the bat population $X = {x_1,x_2,...,x_n}$\;
		\For{each bat $x_i$ in the population}{
				Initialize the pulse rate $r_i$, velocity $v_i$ and loudness $A_i$\;
				Define the pulse frequency $f_i$ at $x_i$\;
		}
		\Repeat{termination criterion not reached}{
				\For{each bat $x_i$ in the population}{
						Generate new solutions through Equations \ref{eq1}, \ref{eq2} and \ref{eq3}\;
						\If{rand$>$$r_i$}{
							Select one solution among the best ones\;
							Generate a local solution around the best one\;
						}
						\If{rand$<$$A_i$ and $f(x_i)$$<$$f(x_*)$}{
							Accept the new solution\;
							Increase $r_i$ and reduce $A_i$\;
						}
				}
		}
		Rank the bats and return the current best bat of the population\;
   \caption{Pseudocode of the basic BA}
	 \label{alg:BA}
\end{algorithm}

After these initialization steps, the algorithm starts its main phase. For each generation, every bat of the swarm moves by updating its velocity and position. For this movement, the following equations are used:

\begin{equation}
	f_i = f_{min} + (f_{min}-f_{max})\beta \label{eq1}
\end{equation}
\begin{equation}
	v_i^t = v_i^{t-1} + [x_i^{t-1} - x_*]f_i \label{eq2}
\end{equation}
\begin{equation}
	x_i^t = x_i^{t-1} + v_i^t \label{eq3}
\end{equation}
where the parameter $\beta$ is a randomly generated number in the [0,1] interval. Additionally, $x_*$ denotes the current best solution in the swarm, and $v_i^t$ and $x_i^t$ represent the velocity and position of a bat $i$ at time step $t$. Finally, the results of Equation (\ref{eq1}) is used to control the range and pace of bats movement. In addition, for the local search part, whether a solution is selected among the best ones, a new solution for each bat is generated using a random walk

\begin{equation}
	x_{new} = x_{old} + \epsilon A^t \label{eq4}
\end{equation}
where $\epsilon$ is a randomly generated number within the interval [-1,1], and $A^t$ is the average loudness of the swarm at time step $t$. Finally, the loudness $A_i$ and the rate $r_i$ of each bat have to be updated if the conditions shown in the line 14 of Algorithm \ref{alg:BA} are met. This update is conducted as follows:

\begin{equation}
	A_i^{t+1} =  \alpha A_i^{t+1} \label{eq5}
\end{equation}
\begin{equation}
	r_i^{t+1} =  r_i^0 [1-\exp(-\gamma t)] \label{eq6}
\end{equation}
where $\alpha$ and $\gamma$ are constants. Thereby, for any 0$<$$\alpha$$<$1 and $\gamma$$>$0 we have

\begin{equation}
	a^t_i\rightarrow 0, r_i^t\rightarrow r_i^0, \ as \ t\rightarrow \infty \label{eq7}
\end{equation}

In many studies of the literature, $\alpha$ = $\gamma$ is used in order to simplify the implementation of the algorithm. Specifically, we use $\alpha$ = $\gamma$ = 0.98 in this work. We have chosen this value empirically using a [0.90, 0.99] range.

\section{The Traveling Salesman Problem}
\label{sec:TSP}

The TSP and ATSP are two of the most well-known and widely studied problems throughout history in computer science and operations research. As many other combinatorial optimization and routing problems, both problems are considered NP-Hard. This is the main reason of their great scientific interest. Thereby, the TSP and all its variants are used in a great number of research works every year as benchmarking problems (\cite{urrutia2015dynamic,wang2015approximate}). The TSP and ATSP can be defined as a complete graph $G= (V,A)$, where $V= \{v_1,v_2,\dots,v_n\}$ is the set of vertices which represents the nodes of the system, and $A= \{(v_i,v_j): v_i,v_j  \in V,i\neq j \}$ is the set of arcs which represents the interconnection between these nodes. Besides that, each arc has an associated $c_{ij}$ cost. In the symmetric version of the TSP the cost of traveling between two nodes is the same in both directions, i.e., $c_{ij}$ = $c_{ji}$. On the other hand, although there may be arcs where $c_{ij}$ = $c_{ji}$, in general $c_{ij}$ $\neq$ $c_{ji}$ for the ATSP.

The objective of the TSP and ATSP is to find a route that, starting and finishing at the same node, visits every node once and that minimizes the total cost of the path. The objective function for these problems is the total cost of the route.

\begin{figure}[tb]
	\centering
		\includegraphics[width=0.7\textwidth]{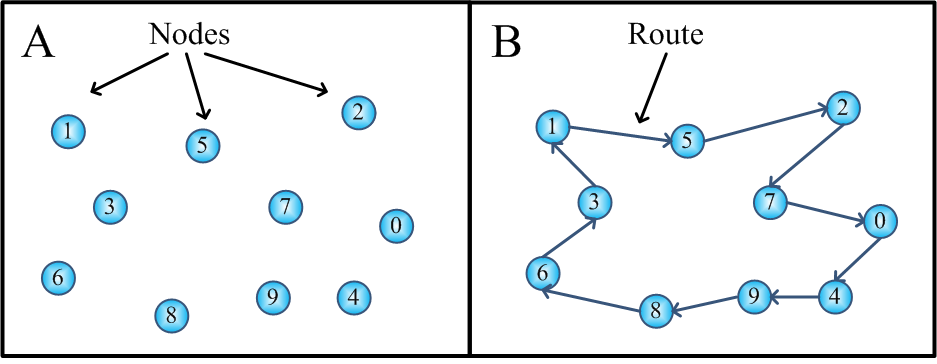}
	\caption{Possible TSP and ATSP 10-noded instance and a feasible solution}
	\label{fig:TSP}
\end{figure}

In this paper, we have used the well-known path representation \\
(\cite{path}) for the encoding of TSP and ATSP solutions. In this way, each solution is encoded as a permutation of numbers, which represents the order in which the nodes are visited. Using as an example a possible 10-node instance of the TSP, or ATSP, one solution would be represented as $X=(1, 5, 2, 7, 0, 4, 9, 8, 6, 3)$. This situation is depicted in Figure \ref{fig:TSP}.

\section{Our Improved Discrete Bat Algorithm for the TSP and ATSP}
\label{sec:OurBat}

In this section, we will explain our adaption of the classic BA to solve the TSP and the ATSP (Section \ref{sec:DBA}). Furthermore, we also explain an improved version of this algorithm (Section \ref{sec:IBA}).

\subsection{Discrete Bat Algorithm for the TSP and ATSP}
\label{sec:DBA}

First of all, it is noteworthy that, as has been said in Section \ref{sec:related}, the original bat algorithm has been applied primarily to continuous optimization problems. As known, both TSP and ATSP are combinatorial optimization problems. Therefore, some modifications of the original BA are needed in order to prepare it for addressing the TSP and ATSP.

In our proposed algorithm, each bat in the swarm represents a possible and feasible solution for the TSP (or ATSP). Additionally, as has been detailed in Section \ref{sec:TSP}, the total traveling cost of the route has been used as the objective function.

Regarding the basic parameters of the classic BA, which are $r_i$, $A_i$, $f_i$ and $v_i$, the philosophy of the first two has remained in exactly the same form. In addition, in order simplify the complexity of the algorithm, the parameter ``frequency", $f_i$ has not been taken into account in our discrete versions of the BA. Finally, the ``velocity", $v_i$, has been modified. In the basic version of the BA this parameter is calculated as has been shown in Equation (\ref{eq2}):

	\[
		v_i^t = v_i^{t-1} + [x_i^{t-1} - x_*]f_i
\]

We can deduce from this formula that the velocity of a bat $i$ at time step $t$ depends on the $v_i$ of the bat $i$ in the previous time step, the difference between the bat $i$ and the best bat in the swarm, and the $f_i$. As can be easily understood, this parameter cannot be used in the same way in our discrete version of the BA. With the intention of adapting the algorithm as accurately as possible, we have considered appropriate to relate $v_i$ with the distance between the bat $i$ and the best bat of the swarm. For this reason, we have adapted $v_i$ using the well-known Hamming Distance in the following way:

\begin{equation}
	v_i^t = \textrm{Random}[1,\textrm{HammingDistance}(x_i^t,x_*)] \label{eq8}
\end{equation}

In other words, the $v_i$ of a bat $i$ at time step $t$ is a random number between 1, and the difference between this bat and the best bat of the swarm. This difference is represented by the Hamming Distance. The Hamming distance between two bats is the number of non-corresponding elements in the sequence. For example, taking into account the following two bats in a hypothetical TSP instance composed by 8 nodes,

\[x_1: [0,1,2,3,4,5,6,7]\]
\[x_2: [0,1,3,2,5,4,6,7]\]
the Hamming Distance between $x_1$ and $x_2$ would be 4.

Furthermore, regarding the generation of new solutions, in the classic BA the movement of the bats is made following the Equation \ref{eq3}:

\[x_i^t = x_i^{t-1} + v_i^t\]

In this case, we can deduce from this formula that the position of a bat $i$ at time step $t$ depends on the $v_i$ of the bat $i$ and its previous position at time step $t-1$. As previously said, this formula cannot be applied directly to the TSP and ATSP in this way. For this reason, we have developed a modification of it. In our study, two well-known successor operators have been used for the movement of the bats:

\begin{itemize}
	\item \textit{2-opt:} This function was defined by Lin in 1965 (\cite{2opt}) and, since then, it has been widely used for solving routing problem (\cite{2optEj1,2optEj2}). The 2-opt eliminates at random two arcs within the existing path and creates two new arcs, avoiding the generation of sub tours.
	\item \textit{3-opt:} The 3-opt operation, proposed also by Lin, is similar to 2-opt, with the difference that in this case the arcs removed are 3. The complexity of using this operator is greater than the 2-opt. Despite this, the operator has been used a large number of times throughout the history (\cite{3opt1,3opt2}).
\end{itemize}

Thereby, the movement performed by each bat $i$ at each time step $t$ is the following:

\begin{equation}
	x_i^t \leftarrow 2-opt(x_i^{t-1},v_i^t) \label{eq9}
\end{equation}

Namely, at each generation, each bat examines a $v_i$ number of its neighbors, and it selects the best one as its current movement. In other words, the bat $i$ performs a $v_i$ number of 2-opt execution and it chooses the best one. In the case of 3-opt, Equation (9) becomes $x_i^t \leftarrow 3-opt(x_i^{t-1},v_i^t)$.

Finally, in relation to local procedure represented in lines 10-12 of Algorithm \ref{alg:BA}, if $rand>r_i$ one solution is selected among the best ones (in our experiments, one bat among the 10 best ones), and a local solution is generated around this one. To generate this local solution, the best neighbor of the chosen bat is selected using also the 2-opt and 3-opt moves.

\subsection{Our proposed improvement for the discrete Bat Algorithm}
\label{sec:IBA}

The BA described in the previous section is the basic discrete BA that we have proposed for solving the TSP and ATSP. In addition, in this paper we also propose a simple but effective improvement in the structure of this basic BA. This improvement is related with the movement behavior of the bats. In the classic version of the BA, all the bats perform their movement following the same pattern throughout the entire execution, regardless of the point in the solution space in which each bat is located.

In our proposed IBA, we have provided some kind of intelligence to all the bats of the swarm. Thereby, each bat moves in a different way depending on its position in relation to the best bat of the swarm. In this way, when one bat is going to perform its movement, it examines its $v_i^t$. If this $v_i^t$ is high (greater than $n$/2, where $n$ is the number of nodes of the TSP-ATSP instance), we can assume that it is far from the best bat of the swarm, and we can conclude that it needs a \textit{large move}. In the other case, if $v_i^t$$<$$n$/2, we can think that the bat is in a promising point of the solution space. Therefore, this bat will perform a \textit{short move}. In our proposal, we have used the 2-opt for \textit{short moves}, and the 3-opt as \textit{large moves}.

This simple modification allows the population individuals to crawl the solution space using different neighborhood structures along the execution. This fact considerably enhances the exploration capacity of the technique, leading to an improvement in the results quality. Finally, the pseudocode of the proposed IBA is depicted in Algorithm \ref{alg:IBA}

\begin{algorithm}[tb]
	 \SetAlgoLined
		Define the objective function $f(x)$\;
		Initialize the bat population $X = {x_1,x_2,...,x_n}$\;
		\For{each bat $x_i$ in the population}{
				Initialize the pulse rate $r_i$, velocity $v_i$ and loudness $A_i$\;
		}
		\Repeat{termination criterion not reached}{
				\For{each bat $x_i$ in the population}{
						Generate new solution\;
						\eIf{$v_i^t$$<$n/2}{
								$x_i \leftarrow 2-opt(x_i^{t-1},v_i^t$)\;
							}{
								$x_i \leftarrow 3-opt(x_i^{t-1},v_i^t$)\;
							}
						\If{rand$>$$r_i$}{
							Select one solution among the best ones\;
							Generate a new bat selecting the best neighbor around the chosen bat using the 2-opt or the 3-opt\;
						}
						\If{rand$<$$A_i$ and $f(x_i)$$<$$f(x_*)$}{
							Accept the new solution\;
							Increase $r_i$ and reduce $A_i$\;
						}
				}
		}
		Rank the bats and return the current best bat of the population\;
   \caption{Pseudocode of the proposed IBA. n = number of nodes of the instance.}
	 \label{alg:IBA}
\end{algorithm}

\section{Experimentation}
\label{sec:exp}

In this section the experimentation performed in this study is detailed. First of all, in Section \ref{sec:expIBA}, a qualitative comparison between the proposed IBA and the presented discrete BA is depicted. Then, in Section \ref{sec:expIBA2}, the results obtained by the IBA are shown and compared with the ones obtained by the other five alternatives. Finally, in Section \ref{sec:est}, the statistical analysis of these results and the convergence behaviour analysis are shown. All the tests conducted in this work have been performed on an Intel Core i5 – 2410 laptop, with 2.30 GHz and a RAM of 4 GB. Java has been used as the programming language. For the TSP, 22 instances have been used, and they have been obtained from the TSPLIB Benchmark (\cite{TSPLib}). In addition, for the ATSP 15 instances have been chosen, obtained from the same benchmark. Overall, 37 instances have been used with 17 to 1002 nodes. Every instance has been run 20 times, and each one has a number in its name which represents the number of nodes it has.

\subsection{Experimentation between presented discrete BA and the proposed IBA}
\label{sec:expIBA}

As has been mentioned in the introduction of this section, an experimentation has been performed in order to prove that the proposed IBA performs better than the basic version of the BA. In this experimentation, the results obtained by the IBA for 35 TSP-ATSP instances have been compared with the ones obtained by two different versions of the basic BA. These results are shown in Table \ref{tab:ResultsTSP}. In this Table, the results average, best solution found, standard deviation and average runtime (in seconds) are shown. In order to facilitate the replicability of this work, the parametrizations used for these three approaches are summarized in Table \ref{tab:Parametrization1}. It is important to highlight that the initial population of bats is randomly generated. In addition, as termination criterion, every execution finishes when there are $n+\sum_{k=1}^n{k}$ generations without improvements in the best solution, where $n$ is the size of the problem.

\begin{table}[tb]
	\centering
	\setlength{\tabcolsep}{2pt}
	\scalebox{0.6}{
		\begin{tabular}{| l | l || l | l || l | l |}
		  \hline
		  \multicolumn{2}{|c||}{IBA} & \multicolumn{2}{c||}{BA1} & \multicolumn{2}{c|}{BA2}\\
			\hline
			\hline Parameter & Value & Parameter & Value & Parameter & Value \\
			\hline
			Population size & 50 & Population size & 50 & Population size & 50 \\
			Movement functions & 2-opt \& 3-opt & Movement function & 2-opt & Movement function & 3-opt\\
			Initial $A_i^0$ & Random number in [0.7,1.0] & Initial $A_i^0$ & Random number in [0.7,1.0] & Initial $A_i^0$ & Random number in [0.7,1.0]  \\
			Initial $r_i^0$ & Random number in [0.0,0.4] & Initial $r_i^0$ & Random number in [0.0,0.4] & Initial $r_i^0$ & Random number in [0.0,0.4]  \\
			$\alpha$ \& $\gamma$ & 0.98 & $\alpha$ \& $\gamma$ & 0.98 & $\alpha$ \& $\gamma$ & 0.98 \\
			\hline
			
			\hline
		\end{tabular}
	}
	\caption{Parametrization of the IBA, BA1 and BA2 for the TSP and ATSP.}
	\label{tab:Parametrization1}
\end{table}

In addition, in order to determine if IBA average is significantly different than the averages obtained by other techniques, we have performed Student’s $t$-test. The $t$ statistic has the following form:

\[t = \frac{\overline{X_1} - \overline{X_2}}{\sqrt{\frac{(n_1 -1)SD^2_1 + (n_2 -1)SD^2_2}{n_1 + n_2 -2}} \frac{n_1 + n_2}{n_1 n_2}}\]

where:

\

$\overline{X_1}$: Average of $IBA$

$SD_1$: Standard deviation of $IBA$,

$\overline{X_2}$: Average of the other technique,

$SD_2$: Standard deviation of the other technique,

$n_1$: $IBA$ size,

$n_2$: Size of the other technique,

\

In Table \ref{tab:ResultsTSP}, we show a direct comparison between IBA and each of the other techniques using the Student's $t$-test. The $t$ values shown can be positive, neutral, or negative. The double positive value (++) of $t$ indicates that IBA is significantly better than the technique with which it is facing. In the opposite case (- -), IBA obtains significant worse solutions. If $t$ is single positive (+), IBA shows to be better but not significantly. On the other hand, if the result is single negative (-), IBA demonstrates to be worse, but not in a significant way. Finally, a neutral value of $t$ depicts equality in the results. We stated confidence interval at the 95\% confidence level ($t_{0.05}$ = 1.96). In this study the numerical value of $t$ is also displayed. Thereby, the difference in results may be seen more easily.

\begin{table}[htb]
	\centering
	\setlength{\tabcolsep}{2pt}
	\scalebox{0.7}{
		\begin{tabular}{| l | c || r r r r || r r r r | c || r r r r | c |}
		  \hline
		  \multicolumn{2}{|c||}{Instance} & \multicolumn{4}{c||}{Improved BA} & \multicolumn{5}{c||}{Basic BA1} & \multicolumn{5}{c|}{Basic BA2}\\
			\hline
			\hline Name & Optima & Avg. & Best & S. dev. & Time & Avg. & Best & S. dev. & Time & t-test & Avg. & Best & S. dev. & Time & t-test\\
			\hline Oliver30 & 420 & \textbf{420.0} & \textbf{420} & 0.0 & 0.4 & \textbf{420.0} & \textbf{420} & 0.0 & 0.3 & * (0.0) & \textbf{420.0} & \textbf{420} & 0.0 & 0.4 & * (0.0)\\
			Eilon50 & 425 & \textbf{427.4} & \textbf{425} & 1.3 & 1.5 & 433.3 & 427 & 3.4 & 1.4  & ++ (7,2) & 438.0 & \textbf{425} & 5.4 & 1.4 & ++ (8.5)\\
			Eil51 & 426 & \textbf{428.1} & \textbf{426} & 1.6 & 1.7 & 438.3 & 430 & 2.5 & 1.7 & ++ (15.3) & 436.8 & 429 & 5.3 & 1.9 & ++ (7.0)\\
			Berlin52 & 7542 & \textbf{7542.0} & \textbf{7542} & 0.0 & 2.1 & 7676.0 & \textbf{7542} & 104.4 & 2.8 & ++ (5.7) & 7681.9 & \textbf{7542} & 112.3 & 2.7 & ++ (5.5)\\
			St70 & 675 & \textbf{679.1} & \textbf{675} & 2.8 & 3.9 & 696.4 & \textbf{675} & 6.3 & 4.1  & ++ (11.2) & 694.3 & \textbf{675} & 9.7 & 4.6 & ++ (6.7)\\
			Eilon75 & 535 & \textbf{547.4} & \textbf{535} & 3.9 & 4.5 & 555.6 & 545 & 7.3 & 4.7  & ++ (4.4) & 562.7 & 550 & 8.9 & 5.1 & ++ (7.0)\\
			Eil76 & 538 & \textbf{548.1} & 539 & 3.8 & 5.1 & 558.8 & \textbf{538} & 9.0 & 5.5 & ++ (4.8) & 560.5 & 540 & 11.6 & 5.8 & ++ (4.5)\\
			KroA100 & 21282 & \textbf{21445.3} & \textbf{21282} & 116.5 & 10.6 & 21884.2 & 21292 & 213.6 & 10.1 & ++ (8.0) & 21989.4 & 21300 & 305.2 & 12.1 & ++ (7.4)\\
			KroB100 & 22140 & \textbf{22506.4} & \textbf{22140} & 221.3 & 11.1 & 22842.9 & 22373 & 231.2 & 12.1 & ++ (4.7) & 22946.7 & 22380 & 291.3 & 12.9 & ++ (5.3)\\
			KroC100 & 20749 & \textbf{21050.0} & \textbf{20749} & 164.7 & 12.0 & 21476.6 & 20802 & 235.1 & 12.0 & ++ (6.6) & 21631.1 & 20802 & 325.0 & 12.8 & ++ (7.1)\\
			KroD100 & 21294 & \textbf{21593.4} & \textbf{21294} & 141.6 & 11.7 & 22001.4 & 21727 & 170.5 & 12.6 & ++ (8.2) & 22053.5 & 21730 & 300.4 & 13.0 & ++ (6.1)\\
			KroE100 & 22068 & \textbf{22349.6} & \textbf{22068} & 169.6 & 11.4 & 22771.5 & 22323 & 216.5 & 12.0 & ++ (6.8) & 22790.2 & 22323 & 284.6 & 12.3 & ++ (5.9)\\
			Eil101  & 629 & \textbf{646.4} & 634 & 4.9 & 13.1 & 667.1 & 640 & 4.4 & 13.5 & ++ (14.0) & 670.0 & 642 & 8.4 & 14.1 & ++ (10.8)\\
			Pr107 & 44303 & \textbf{44793.8} & \textbf{44303} & 232.4 & 12.1 & 45030.4 & 44618 & 184.4 & 14.4 & ++ (3.5) & 45242.1 & 44701 & 259.3 & 15.8 & ++ (5.7)\\
			Pr124 & 59030 & \textbf{59412.1} & \textbf{59030} & 265.9 & 18.5 & 59627.2 & \textbf{59030} & 395.9 & 19.7 & + (1.9) & 59791.0 & 59074 & 448.3 & 20.5 & ++ (3.2)\\
			Pr136 & 96772 & \textbf{99351.2} & 97547 & 707.2 & 23.4 & 101630.5 & 100485 & 732.8 & 24.2 & ++ (10.0) & 101903.6 & 100500 & 893.1 & 25.3 & ++ (10.0)\\
			Pr144 & 58537 & \textbf{58876.2} & \textbf{58537} & 295.6 & 30.3 & 58961.9 & 58588 & 227.4 & 29.9 & + (1.0) & 59012.5 & 58602 & 301.0 & 31.2 & + (1.4)\\
			Pr152 & 73682 & \textbf{74676.9} & 73921 & 426.5 & 31.0 & 74993.9 & 74172 & 429.3 & 28.5 & ++ (2.3) & 75241.0 & 74172 & 539.3 & 30.0 & ++ (3.6)\\
			Pr264 & 49135 & \textbf{50908.3} & 49756 & 887.0 & 92.5 & 52412.4 & 50256 & 995.3 & 90.3 & ++ (16.3) & 52628.2 & 50306 & 1002.3 & 93.7 & ++ (18.5)\\
			Pr299 & 48191 & \textbf{49674.1} & 48310 & 1200.1 & 147.2 & 50434.0 & 49142 & 1528.0 & 150.3 & ++ (5.6) & 50232.6 & 49193 & 1739.1 & 153.2 & ++ (3.8)\\
			\hline
			\hline
			br17 & 39 & \textbf{39.0} & \textbf{39} & 0.0 & 0.2 & \textbf{39.0} & \textbf{39} & 0.0 & 0.4 & * (0.0) & \textbf{39.0} & \textbf{39} & 0.0 & 0.8 & * (0.0) \\
			ftv33 & 1286 & \textbf{1318.1} & \textbf{1286} & 25.7 & 2.2 & 1390.6 & 1348 & 26.8 & 1.8 & ++ (8.7) & 1387.3 & 1334 & 30.4 & 3.1 &  ++ (7.7)\\
			ftv35 & 1473 & \textbf{1493.7} & \textbf{1473} & 8.0 & 2.5 & 1559.6 & 1490 & 41.2 & 1.9 & ++ (7.0) & 1571.9 & 1529 & 31.2 & 3.6 & ++ (10.8) \\
			ftv38 & 1530 & \textbf{1562.0} & \textbf{1530} & 13.79 & 3.1 & 1603.2 & \textbf{1530} & 43.6 & 2.5 & ++ (4.0) & 1657.8 & 1615 & 32.0 & 4.2 & ++ (12.2)\\
			p43 & 5620 & \textbf{5620.0} & \textbf{5620} & 0.0 & 3.0 & 5654.6 & 5632 & 9.1 & 2.7 & ++ (16.9) & 5621.2 & \textbf{5620} & 0.8 & 3.7 & ++ (6.0)\\
			ftv44 & 1613 & \textbf{1683.7} & \textbf{1613} & 27.2 & 5.0 & 1774.5 & 1725 & 43.0 & 4.2 & ++ (7.9) & 1817.4 & 1754 & 55.5 & 6.8 & ++ (9.6)\\
			ftv47 & 1776 & \textbf{1863.6} & 1796 & 39.3 & 4.7 & 1959.3 & 1842 & 62.4 & 3.9 & ++ (5.8) & 2031.6 & 1937 & 48.3 & 6.7 & ++ (12.0)\\
			ry48p & 14422 & \textbf{14544.8} & \textbf{14422} & 79.7 & 4.2 & 15172.9 & 14790 & 153.7 & 3.7 & ++ (16.2) & 15069.7 & 14798 & 150.4 & 6.9 & ++ (13.7)\\
			ft53 & 6905 & \textbf{7294.1} & 7001 & 196.9 & 6.5 & 7732.4 & 7105 & 252.6 & 5.1 & ++ (6.1) & 7910.9 & 7452 & 151.8 & 8.2 & ++ (11.0)\\
			ftv55 & 1608 & \textbf{1737.5} & \textbf{1608} & 50.5 & 6.9 & 1861.9 & 1686 & 81.1 & 5.3 & ++ (5.8) & 1895.0 & 1806 & 45.7 & 8.3 & ++ (10.3)\\
			ftv64 & 1839 & \textbf{1999.2} & 1879 & 68.2 & 7.2 & 2215.8 & 2068 & 74.8 & 6.2 & ++ (9.5) & 2294.2 & 2122 & 58.1 & 9.0 & ++ (14.8)\\
			ftv70 & 1950 & \textbf{2233.2} & 2111 & 48.8 & 8.1 & 2434.9 & 2238 & 84.3 & 6.7 & ++ (9.2) & 2507.8 & 2314 & 110.8 & 10.5 & ++ (10.1)\\
			ft70 & 38673 & \textbf{40309.7} & 39901 & 237.2 & 8.2 & 2434.9 & 2238 & 84.3 & 6.7 & ++ (6.3) & 42506.3 & 42070 & 260.4 & 10.7 & ++ (27.8)\\
			kro124p & 36230 & \textbf{39213.7} & 37538 & 947.5 & 15.4 & 41772.6 & 40070 & 1872.2 & 13.2 & ++ (5.4) & 43200.7 & 42307 & 611.7 & 18.7 & ++ (15.8)\\
			rbg323 & 1326 & \textbf{1640.9} & 1615 & 30.4 & 243.6 & 1738.4 & 1713 & 16.5 & 237.4 & ++ (14.8) & 1828.0 & 1713 & 92.7 & 251.7 & ++ (8.3)\\
			\hline
		\end{tabular}
	}
	\caption{Results of the proposed IBA and two different Basic BAs for the TSP and ATSP.}
	\label{tab:ResultsTSP}
\end{table}

As can be concluded viewing the results shown in Table \ref{tab:ResultsTSP}, the IBA meets or outperforms the outcomes obtained by both basic discrete versions of the BA in the 100\% of the cases. In addition, taking into account the performed Student's $t$-test, the differences in the results are significant in the 90\% of the cases (63 out of 70 confrontations). The reason why the IBA is a better technique can be explained as follows: the bats that compose the population of the IBA have the option of exploring different neighborhood structures. This fact occurs because bats can switch their movement function method throughout the execution of the algorithm depending on the value of their $v_i$ parameter. As we have explained in other studies (\cite{GB}), this change of neighborhood structure is an efficient mechanism to avoid local optima in routing problems, and it helps bats to explore the solution space in different ways.

\subsection{Experimentation between the proposed IBA and the literature techniques}
\label{sec:expIBA2}

For sake of clarity, and before starting with the details of the performed experiments, we now briefly resume the basic principles of each technique which has been used in the experimentation.

Regarding the first of these techniques, the SA, it is one of the most popular local search techniques. It is based on the physical principle of cooling metal. Using that analogy, a SA generates an initial solution and the process proceeds by selecting new solutions randomly. This new solutions are not always better than the current ones, but they can be accepted probabilistically. Furthermore, as time passes and the temperature decreases (the metal becomes stronger), the probabilty of accepting worse solutions decreases, until it finally reaches 0. With the aim of performing a fair and rigorous experimentation, and taking into account that the IBA is a population technique, we have used a distributed version for the SA: the ESA (\cite{ESA}).

On the other hand, GAs are one of the most successful meta-heuristics for solving combinatorial optimization problems. Thanks to their easy application and good performance, GAs have been used to solve many complex problems framed in various fields. GAs were proposed in 1975 by Holland (\cite{genetic3}), in an attempt to imitate the genetic process of living organisms, and the law of the evolution of species. Anyway, their practical use to solve complex optimization problems was shown later, by Goldberg (\cite{genetic1}) and De Jong (\cite{genetic2}). With the aim of overcoming the drawbacks of GAs, such as premature convergence to a local optimum, and the imbalance between exploration and exploitation, Parallel GAs were proposed (PGA). PGAs are particularly easy to implement and promise substantial gains in performance. Reviewing the literature it can be seen that there are different ways to parallelize GA. The generally used classification divides parallel GAs in three categories: Fine Grain, Panmitic model and Island model. This last category is the most used, and it consists in a multiple populations that evolve separately most of the time and exchange individuals occasionally. This is the approach employed for the IDGA developed in our study.

Regarding the Imperialist Competitive Algorithm, it was proposed by Atashpaz-Gargari and Lucas (\cite{imperialist}), and it is based on the concept of imperialism. The ICA divides the population into various empires, which evolve independently. Individuals of the population are called countries, and they are divided into two types: imperialist states (best country of the empire) and colonies. In this technique, the colonies make ​​their movement through the solution space basing on the imperialist state. Meanwhile, empires compete between them, trying to conquer the weakest colonies of each other. This way, powerless empires could collapse and disappear, dividing their colonies among other empires.

The last used technique is a DFA. The first version of a Firefly Algorithm was proposed by Xin-She Yang in 2008. This nature-inspired algorithm is based on the flashing behaviour of fireflies, which acts as a signal system to attract other fireflies. As can be seen in several surveys (\cite{FS2,FS3}), the FA has been successfully applied to many different optimization fields and problems since its proposal. In addition, it still attracts a lot of interests in the current scientific community (\cite{ma2015navigability,liang2015enhanced,singh2015disk}).

\begin{table}[tb]
	\centering
	\setlength{\tabcolsep}{2pt}
	\scalebox{0.7}{
		\begin{tabular}{| l | l || l | l || l | l |}
		  \hline
		  \multicolumn{2}{|c||}{ESA} & \multicolumn{2}{c||}{GA} & \multicolumn{2}{c|}{IDGA}\\
			\hline
			\hline Parameter & Value & Parameter & Value & Parameter & Value \\
			\hline
			Population size & 50 & Population size & 50 & Population size & 4 subpob. of 13 individuals \\
			Successor functions & 2-opt \& 3-opt & Crossover function & OX & Crossover functions & OX \& OBX \\
			Temperature & $-sup \Delta f/ ln(p) $ & Mutation functions & 2-opt \& 3-opt & Mutation functions & 2-opt \& 3-opt \\
			Cooling constant & 0.95 & Cross. prob. & 0.95  & Cross. prob. & 0.95, 0.9, 0.8 \& 0.75 \\
			& & Cross. prob. & 0.25 & Mut. prob. & 0.05, 0.1, 0.2 \& 0.25 \\
		  & & Selection func. & Binary tournament  & Selection func. & Binary tournament   \\
			& & Survivor func. & Binary tournament & Survivor func. & Binary tournament \\
			& &  &  & Migration Strat. & Best-Replace-Worst (\cite{migration1}) \\
			\hline
			
			\hline
		\end{tabular}
	}
	\caption{Parametrization of the ESA, GA and IDGA for the TSP and ATSP. OX: Order Crossover (\cite{OX}). OBX: Order Based Crossover (\cite{OBX}). $-sup \Delta f$ is the difference in the objective function of the best and the worse individuals of the initial population, and $p$=0.95.}
	\label{tab:Parametrization2}
\end{table}

As has been mentioned previous sections, the TSP is one of the standard test problems used in performance analysis of discrete optimization algorithms. In this way, even though the IBA obtains good results for both TSP and ATSP (it reaches the optimal solution in 14 out of 22 instance for the TSP and in 8 out of 15 for the ATSP, and in average, its solutions deviate a 1.38\% from the optimal for the TSP and in 6,3\% for the ATSP), it is important to highlight that the main goal of this study is not to find an optimal solution to these problems. Instead, we use both problems as benchmarking problems, which means that the principal objective of this research is to prove that the BA can be easily adapted to routing problems, and that the IBA is a promising approximation method for solving the TSP and ATSP. To reach this objective, we prove that the IBA can outperform some of the most used and well-known metaheuristics of the literature using classical non-heuristic functions. Thereby, for the results comparison we have chosen three of the most historically famous and successful techniques, the GA, the SA and the IDGA, and two recently proposed techniques, the DFA and the DICA.

It is important to highlight that, as far as possible, we have been used the same operators in similar parameters for all the algorithms implemented for the experimentation. In this way, our aim is to conclude which algorithm obtains better results using similar operators similar number of times. Furthermore, with the intention of facilitating the replicability of this study, we also show in Table \ref{tab:Parametrization2} the parametrization used for these three algorithms. It is worth pointing out that all the individual are randomly generated. Besides this, we can find two successor functions for the ESA, which means that every individual has its own randomly assigned successor function. A similar procedure has been used in the IDGA with the crossover function, and in the GA and IDGA with the mutation function. Finally, as for the termination criterion, it is the same as for the IBA.

On the other hand, the parametrization used for IBA is the same shown in the previous section in Table \ref{tab:Parametrization1}. Additionally, the results obtained by the IBA, ESA, GA and IDGA techniques for the 37 instances are depicted in Table \ref{tab:ResultsTSP2}. In this table, we have shown the results average, best results, standard deviation and average runtime (in seconds). In addition, we have represented bolded the best results average, and the best solution found only whether it is the optimal one. As can be seen, and with the intention of not duplicating experiments, the results shown in Table \ref{tab:ResultsTSP2} for the IBA are the same as have been depicted in Table \ref{tab:ResultsTSP}.

\begin{table}[tb]
	\centering
	\setlength{\tabcolsep}{2pt}
	\scalebox{0.7}{
		\begin{tabular}{| l | c || r r r r | r r r r | r r r r | r r r r |}
		  \hline
		  \multicolumn{2}{|c||}{Instance} & \multicolumn{4}{c|}{IBA} & \multicolumn{4}{c|}{ESA} & \multicolumn{4}{c|}{GA} & \multicolumn{4}{c|}{IDGA}\\
			\hline
			\hline Name & Optima & Avg. & Best & S. dev. & Time & Avg. & Best & S. dev. & Time & Avg. & Best & S. dev. & Time & Avg. & Best & S. dev. & Time \\
			\hline Oliver30 & 420 & \textbf{420.0} & \textbf{420} & 0.0 & 0.4 & \textbf{420.0} & \textbf{420} & 0.0 & 0.7 & 422.8 & \textbf{420} & 3.4 & 0.2 & 421.5 & \textbf{420} & 2.1 & 0.2\\
			Eilon50 & 425 & 427.4 & \textbf{425} & 1.3 & 1.5 & 429.0 & 427 & 1.7 & 2.2 & 427.6 & 426 & 5.8 & 1.2 & \textbf{427.0} & \textbf{425} & 2.2 & 0.7\\
			Eil51 & 426 & \textbf{428.1} & \textbf{426} & 1.6 & 1.7 & 431.6 & \textbf{426} & 2.9 & 2.1 & 440.8 & 427 & 7.3 & 1.7 & 434.4 & \textbf{426} & 4.5 & 1.2\\
			Berlin52 & 7542 & \textbf{7542.0} & \textbf{7542} & 0.0 & 2.1 & \textbf{7542.0} & \textbf{7542} & 0.0 & 2.3 & \textbf{7542.0} & \textbf{7542} & 0.0 & 2.4 & \textbf{7542.0} & \textbf{7542} & 0.0 & 2.3\\
			St70 & 675 & \textbf{679.1} & \textbf{675} & 2.8 & 3.9 & 682.1 & \textbf{675} & 3.9 & 4.5 & 709.8 & \textbf{675} & 5.7 & 4.2 & 690.2 & \textbf{675} & 9.8 & 4.1\\
			Eilon75 & 535 & \textbf{547.4} & \textbf{535} & 3.9 & 4.5 & 550.2 & 545 & 3.9 & 5.4 & 565.6 & 550 & 14.2 & 5.6 & 552.4 & 544 & 7.6 & 4.4\\
			Eil76 & 538 & \textbf{548.1} & 539 & 3.8 & 5.1 & 553.7 & 546 & 4.2 & 5.8 & 565.4 & 545 & 9.8 & 5.6 & 557.7 & 545 & 6.8 & 5.1\\
			KroA100 & 21282 & \textbf{21445.3} & \textbf{21282} & 116.5 & 10.6 & 21481.7 & \textbf{21282} & 150.1 & 14.0 & 21812.4 & 21350 & 420.8 & 9.9 & 21731.8 & 21345 & 340.7 & 10.7\\
			KroB100 & 22140 & \textbf{22506.4} & \textbf{22140} & 221.3 & 11.1 & 22602.2 & 22202 & 210.2 & 13.6 & 22687.4 & 22176 & 407.7 & 10.7 & 22712.6 & 22208 & 312.8 & 10.7\\
			KroC100 & 20749 & \textbf{21050.0} & \textbf{20749} & 164.7 & 12.0 & 21170.4 & \textbf{20749} & 188.7 & 15.4 & 21510.4 & 20861 & 390.2 & 10.2 & 21298.7 & 20830 & 290.7 & 11.2\\
			KroD100 & 21294 & \textbf{21593.4} & \textbf{21294} & 141.6 & 11.7 & 21726.5 & 21500 & 156.9 & 15.9 & 22184.6 & 21492 & 405.0 & 9.7 & 21696.9 & 21582 & 408.9 & 12.1\\
			KroE100 & 22068 & \textbf{22349.6} & \textbf{22068} & 169.6 & 11.4 & 22499.7 & 22099 & 171.4 & 15.0 & 22741.3 & 22150 & 306.0 & 9.4 & 22721.9 & 22110 & 368.0 & 12.6\\
			Eil101  & 629 & \textbf{646.4} & 634 & 4.9 & 13.1 & 658.4 & 650 & 4.4 & 16.3 & 673.8 & 655 & 12.5 & 10.6 & 660.7 & 650 & 7.5 & 11.7\\
			Pr107 & 44303 & \textbf{44793.8} & \textbf{44303} & 232.4 & 12.1 & 44821.5 & 44413 & 179.3 & 16.7 & 45619.6 & 44392 & 1395.4 & 10.8 & 44902.5 & 44428 & 660.3 & 12.91\\
			Pr124 & 59030 & \textbf{59412.1} & \textbf{59030} & 265.9 & 18.5 & 59593.6 & \textbf{59030} & 367.8 & 23.1 & 59901.0 & \textbf{59030} & 562.6 & 17.3 & 59912.8 & 59072 & 532.1 & 17.8\\
			Pr136 & 96772 & \textbf{99351.2} & 97547 & 707.2 & 23.4 & 99858.3 & 98499 & 655.7 & 29.5 & 100472.4 & 98432 & 1225.6 & 23.8 & 99932.7 & 98532 & 1301.2 & 23.7\\
			Pr144 & 58537 & 58876.2 & \textbf{58537} & 295.6 & 30.3 & \textbf{58807.3} & 58574 & 220.9 & 33.9 & 60591.4 & 58599 & 2342.8 & 32.8 & 58893.0 & 58581 & 1012.4 & 32.5\\
			Pr152 & 73682 & \textbf{74676.9} & 73921 & 426.5 & 31.0 & 74969.5 & 74172 & 498.9 & 39.5 & 75658.3 & 74520 & 910.8 & 33.4 & 75126.7 & 74249 & 1005.7 & 32.0\\
			Pr264 & 49135 & \textbf{50908.3} & 49756 & 887.0 & 92.5 & 52198.5 & 51603 & 426.1 & 102.5 & 52499.8 & 51712 & 932.4 & 92.1 & 52290.0 & 51653 & 782.7 & 94.5\\
			Pr299 & 48191 & \textbf{49674.1} & 48310 & 1200.1 & 147.2 & 50532.3 & 49242 & 915.8 & 158.7 & 50817.1 & 49659 & 1585.7 & 147.6 & 50513.3 & 49572 & 1257.9 & 149.94\\
			
			Pr439 & 107217 & \textbf{115256.4} & 11153 & 3825.8 & 201.9 & 116706.9 & 113497 & 4168.4 & 206.4 & 116943.4 & 113576 & 4642.4 & 208.4 & 116436.1 & 113207 & 4513.6 & 205.7 \\
			Pr1002 & 259047 & \textbf{274419.7} & 270016 & 3617.8 & 681.7 & 279419.7 & 273496 & 5120.3 & 683.1 & 279384.7 & 273001 & 5534.4 & 689.4 & 278951.4 & 272893 & 5617.4 & 687.1\\
			
			\hline
			\hline
			br17 & 39 & \textbf{39.0} & \textbf{39} & 0.0 & 0.2 & \textbf{39.0} & \textbf{39} & 0.0 & 0.1 & \textbf{39.0} & \textbf{39} & 0.0 & 0.1 & \textbf{39.0} & \textbf{39} & 0.0 & 0.1 \\
			ftv33 & 1286 & \textbf{1318.1} & \textbf{1286} & 25.7 & 2.2 & 1322.5 & \textbf{1286} & 24.5 & 2.6 & 1409.4 & 1290 & 81.2 & 1.7 & 1402.7 & \textbf{1286} & 99.7 & 2.0\\
			ftv35 & 1473 & 1493.7 & \textbf{1473} & 8.0 & 2.5 & \textbf{1490.3} & \textbf{1473} & 29.5 & 2.5 & 1597.2 & 1490 & 78.4 & 2.0 & 1589.0 & 1498 & 82.1 & 2.3 \\
			ftv38 & 1530 & \textbf{1562.0} & \textbf{1530} & 13.7 & 3.1 & 1568.8 & \textbf{1530} & 21.0 & 2.9 & 1670.4 & 1565 & 67.4 & 2.6 & 1650.1 & 1560 & 72.1 & 2.7 \\
			p43 & 5620 & \textbf{5620.0} & \textbf{5620} & 0.0 & 3.0 &  \textbf{5620.0} & \textbf{5620} & 0.0 & 2.5 & 5625.2 & 5620 & 5.4 & 2.8 & \textbf{5620.0} & 5620 & 0.0 & 2.6\\
			ftv44 & 1613 & \textbf{1683.7} & \textbf{1613} & 27.2 & 5.0 & 1718.9 & 1645 & 39.2 & 5.3 & 1780.0 & 1649 & 94.7 & 4.6 & 1800.3 & 1645 & 120.7 & 5.3\\
			ftv47 & 1776 & \textbf{1863.6} & 1796 & 39.3 & 4.7 & 1879.8 & 1795 & 52.7 & 5.2 & 1963.1 & 1820 & 89.6 & 4.3 & 1957.4 & 1822 & 118.4 & 4.7\\
			ry48p & 14422 & \textbf{14544.8} & \textbf{14422} & 79.7 & 4.2 & 14598.0 & 14485 & 108.7 & 4.6 & 14992.1 & 14545 & 340.7 & 3.9 & 14892.0 & 14530 & 201.8 & 4.3\\
			ft53 & 6905 & \textbf{7294.1} & 7001 & 196.9 & 6.5 & 7314.7 & 6990 & 157.8 & 6.6 & 7568.4 & 7270 & 358.7 & 6.2 & 7445.2 & 7076 & 430.0 & 6.1\\
			ftv55 & 1608 & \textbf{1737.5} & \textbf{1608} & 50.5 & 6.9 & 1822.6 & 1725 & 70.1 & 7.2 & 1871.1 & 1700 & 132.1 & 5.8 & 1970.8 & 1842 & 120.1 & 6.7 \\
			ftv64 & 1839 & \textbf{1999.2} & 1879 & 68.2 & 7.2 & 2072.3 & 1955 & 65.0 & 7.1 & 2205.7 & 2014 & 127.4 & 6.9 & 2262.1 & 2080 & 152.1 & 6.8 \\
			ftv70 & 1950 & \textbf{2233.2} & 2111 & 48.8 & 8.1 & 2312.6 & 2200 & 67.2 & 8.0 & 2315.8 & 2184 & 140.7 & 7.9 & 2351.7 & 2135 & 134.2 & 7.5 \\
			ft70 & 38673 & \textbf{40309.7} & 39901 & 237.2 & 8.2 & 40551.4 & 39650 & 467.2 & 8.7 & 40400.7 & 39407 & 620.4 & 7.6 & 40672.4 & 39241 & 781.8 & 8.2\\
			kro124p & 36230 & \textbf{39213.7} & 37538 & 947.5 & 15.4 & 42132.0 & 40019 & 1250.7 & 17.7 & 42250.3 & 39265 & 1825.4 & 15.3 & 42101.9 & 39099 & 1072.4 & 15.8 \\			
			rbg323 & 1326 & 1640.9 & 1615 & 30.4 & 243.6 & 1685.0 & 1620 & 72.1 & 246.8 & 1631.5 & 1514 & 77.2 & 238.1 & \textbf{1623.5} & 1510 & 82.2 & 238.7\\
			\hline
		\end{tabular}
	}
	\caption{Results of the proposed IBA and ESA, GA and IDGA for the TSP and ATSP.}
	\label{tab:ResultsTSP2}
\end{table}

The main conclusion that we can draw from this experimentation is that the IBA has been proved to be better than the ESA, GA and IDGA. Overall, the IBA has obtained better results in the 81.81\% of the TSP instances (18 out of 22), being worse only in two instances (Eilon50 and Pr144). On the other hand, IBA has outperformed the other alternatives in the 73.33\% of the ATSP instances, getting worse results only in two cases (ftv35 and rbg323). Technique by technique, the IBA has outperformed the ESA in 83.78\% of the instances (31 out of 37). Furthermore, the IBA has performed better than the GA in 91.89\% of the cases (34 out of 37) regarding the GA, and in 86.48\% in relation to the IDGA (32 out of 37). Finally, it is important to highlight that the IBA has obtained worse results compared with each of the alternatives only in 4 occasions (in Pr144 and ftv35 with the ESA, and Eilon50 and rbg323 with the IDGA). This results are confirmed by the statistical tests shown in Section \ref{sec:est}. Besides this, the IBA has reached the optimal solution in 59.45\% of the instances (22 out of 37), outperforming the other techniques also in this respect.

Another important factor that is worth mentioning is the robustness of the IBA in relation to the other techniques. As can be seen in Table \ref{tab:ResultsTSP2}, the standard deviation of the results obtained by the IBA is lower than the ones presented by the other metaheuristics. This means that the quality of the solutions provided by the IBA move in a narrow range. This characteristic gives robustness and reliability to the algorithm, something that is very important if we want to use our technique in a real environment.

Finally, looking at the runtimes, we can say that the differences are not remarkable. While the IBA shows a slightly better performance in this respect compared with ESA and a similar behavior regarding the DGA, the GA has proved to be the best alternative. Anyway, as has been said, these differences are not remarkable and all the execution times shown by every technique are acceptable.

On the other hand, in Table \ref{tab:ResultsTSP3} the results obtained by the IBA in the 37 instances are compared with the ones obtained by a DFA and a DICA. In order to implement both these algorithms, the guidelines given in (\cite{li2015firefly}) and (\cite{yousefikhoshbakht2013new}) have been followed. As we have done previously, we have used similar parameters and functions in these techniques with the intention of obtaining fair conclusions. In this way, the same number of individuals has been used for the IBA, DFA and DICA, and all these techniques base the movements of these individuals on the Hamming Distance function.

\begin{table}[tb]
	\centering
	\setlength{\tabcolsep}{2pt}
	\scalebox{0.7}{
		\begin{tabular}{| l | c || r r r r | r r r r | r r r r |}
		  \hline
		  \multicolumn{2}{|c||}{Instance} & \multicolumn{4}{c|}{IBA} & \multicolumn{4}{c|}{DFA} & \multicolumn{4}{c|}{DICA} \\
			\hline
			\hline Name & Optima & Avg. & Best & S. dev. & Time & Avg. & Best & S. dev. & Time & Avg. & Best & S. dev. & Time \\
			\hline
			Oliver30 & 420 & \textbf{420.0} & \textbf{420} & 0.0 & 0.4 & \textbf{420.0} & \textbf{420} & 0.0 & 0.4 & \textbf{420.}0 & \textbf{420} & 0.0 & 0.5\\
			Eilon50 & 425 & 427.4 & \textbf{425} & 1.3 & 1.5 & \textbf{427.2} & \textbf{425} & 1.8 & 1.3 & 427.9 & \textbf{425} & 2.1 & 1.4\\
			Eil51 & 426 & \textbf{428.1} & \textbf{426} & 1.6 & 1.7 & 430.8 & \textbf{426} & 2.3 & 1.6 & 432.3 & 426 & 3.1 & 1.8\\
			Berlin52 & 7542 & \textbf{7542.0} & \textbf{7542} & 0.0 & 2.1 & \textbf{7542.0} & \textbf{7542} & 0.0 & 2.2 & \textbf{7542.0} & \textbf{7542} & 0.0 & 2.5\\
			St70 & 675 & \textbf{679.1} & \textbf{675} & 2.8 & 3.9 & 685.3 & \textbf{675} & 4.0 & 4.3 & 684.7 & \textbf{675} & 3.7 & 4.1\\
			Eilon75 & 535 & 547.4 & \textbf{535} & 3.9 & 4.5 & \textbf{543.6} & \textbf{535} & 5.3 & 4.6 & 551.7 & 537 & 6.8 & 5.6\\
			Eil76 & 538 & \textbf{548.1} & 539 & 3.8 & 5.1 & 556.8 & 543 & 4.9 & 5.3 & 557.6 & 544 & 5.8 & 5.3 \\
			KroA100 & 21282 & \textbf{21445.3} & \textbf{21282} & 116.5 & 10.6 & 21483.6 & \textbf{21282} & 163.7 & 10.3 & 21500.3 & 21282 & 183.4 & 10.8 \\
			KroB100 & 22140 & \textbf{22506.4} & \textbf{22140} & 221.3 & 11.1 & 22604.8 & 22183 & 243.9 & 11.6 & 22599.7 & 22180 & 244.9 & 11.3\\
			KroC100 & 20749 & \textbf{21050.0} & \textbf{20749} & 164.7 & 12.0 & 21096.3 & 20756 & 148.3 & 12.8 & 21103.9 & 20756 & 161.1 & 11.7\\
			KroD100 & 21294 & \textbf{21593.4} & \textbf{21294} & 141.6 & 11.7 & 21683.8 & 21408 & 163.7 & 12.4 & 21666.8 & 21399 & 174.0 & 12.6\\
			KroE100 & 22068 & \textbf{22349.6} & \textbf{22068} & 169.6 & 11.4 & 22413.0 & 22079 & 183.0 & 11.6 & 22453.3 & 22083 & 196.9 & 11.7\\
			Eil101  & 629 & \textbf{646.4} & 634 & 4.9 & 13.1 & 659.0 & 643 & 8.1 & 13.3 & 663.8 & 644 & 9.6 & 12.0\\
			Pr107 & 44303 & 44793.8 & \textbf{44303} & 232.4 & 12.1 & \textbf{44790.4} & \textbf{44303} & 227.3 & 12.6 & 44803.3 & \textbf{44303} & 302.7 & 12.9\\
			Pr124 & 59030 & 59412.1 & \textbf{59030} & 265.9 & 18.5 & \textbf{59404.3} & \textbf{59030} & 257.9 & 18.8 & 59436.9 & \textbf{59030} & 299.4 & 19.0\\
			Pr136 & 96772 & \textbf{99351.2} & 97547 & 707.2 & 23.4 & 99683.7 & 97716 & 831.3 & 24.1 & 99583.7 & 97736 & 848.9 & 24.0\\
			Pr144 & 58537 & \textbf{58876.2} & \textbf{58537} & 295.6 & 30.3 & 58993.3 & 58546 & 300.1 & 30.9 & 59070.9 & 58563 & 323.0 & 30.7\\
			Pr152 & 73682 & \textbf{74676.9} & 73921 & 426.5 & 31.0 & 74934.3 & 74033 & 483.7 & 32.1 & 74886.7 & 74052 & 513.9 & 32.0\\
			Pr264 & 49135 & \textbf{50908.3} & 49756 & 887.0 & 92.5 & 51837.0 & 50491 & 760.8 & 93.0 & 51943.6 & 50553 & 863.7 & 94.1\\
			Pr299 & 48191 & \textbf{49674.1} & 48310 & 1200.1 & 147.2 & 49839.7 & 48579 & 1305.4 & 149.1 & 49880.3 & 48600 & 1413.7 & 150.3 \\
			
			Pr439 & 107217 & \textbf{115256.4} & 111538 & 3825.8 & 201.9 & 115558.2 & 111967 & 4009.1 & 202.4 & 115763.1 & 111983 & 4219.6 & 203.7\\
			Pr1002 & 259047 & \textbf{274419.7} & 270016 & 3617.8 & 681.7 & 277344.7 & 272003 & 4731.6 & 682.0 & 277308.1 & 272082 & 4293.7 & 684.6\\
			
			\hline
			\hline
			br17 & 39 & \textbf{39.0} & \textbf{39} & 0.0 & 0.2 & \textbf{39.0} & \textbf{39} & 0.0 & 0.2 & \textbf{39.0} & \textbf{39} & 0.0 & 0.3\\
			ftv33 & 1286 & \textbf{1318.1} & \textbf{1286} & 25.7 & 2.2 & 1320.9 & \textbf{1286} & 21.9 & 2.8 & 1324.6 & \textbf{1286} & 28.3 & 2.9\\
			ftv35 & 1473 & 1493.7 & \textbf{1473} & 8.0 & 2.5 & 1498.8 & \textbf{1473} & 10.4 & 2.7 & \textbf{1490.6} & \textbf{1473} & 11.9 & 2.8\\
			ftv38 & 1530 & 1562.0 & \textbf{1530} & 13.7 & 3.1 & \textbf{1560.4} &\textbf{1530} & 14.6 & 3.0 & 1565.6 & 1530 & 15.8 & 3.2 \\
			p43 & 5620 & \textbf{5620.0} & \textbf{5620} & 0.0 & 3.0 & \textbf{5620.0} & \textbf{5620} & 0.0 & 2.8 & \textbf{5620.0} & \textbf{5620} & 0.0 & 3.1\\
			ftv44 & 1613 & \textbf{1683.7} & \textbf{1613} & 27.2 & 5.0 & 1690.8 & 1620 & 32.3 & 5.1 & 1694.3 & 1622 & 54.0 & 5.3\\
			ftv47 & 1776 & 1863.6 & 1796 & 39.3 & 4.7 & \textbf{1858.3} & 1795 & 63.4 & 5.5 & 1873.0 & 1799 & 70.1 & 5.8\\
			ry48p & 14422 & \textbf{14544.8} & \textbf{14422} & 79.7 & 4.2 & 14694.4 & 14453 & 94.7 & 4.4 & 14689.8 & 14463 & 73.6 & 5.4 \\
			ft53 & 6905 & \textbf{7294.1} & 7001 & 196.9 & 6.5 & 7302.0 & 6993 & 186.4 & 6.8 & 7320.1 & 7002 & 200.3 & 6.9\\
			ftv55 & 1608 & \textbf{1737.5} & \textbf{1608} & 50.5 & 6.9 & 1790.6 & 1628 & 64.0 & 7.0 & 1801.4 & 1630 & 83.0 & 7.2 \\
			ftv64 & 1839 & \textbf{1999.2} & 1879 & 68.2 & 7.2 & 2041.6 & 1903 & 73.4 & 7.0 & 2040.8 & 1900 & 81.3 & 7.3\\
			ftv70 & 1950 & \textbf{2233.2} & 2111 & 48.8 & 8.1 & 2290.8 & 2173 & 60.0 & 7.8 & 2322.6 & 2167 & 63.3 & 8.3\\
			ft70 & 38673 & \textbf{40309.7} & 39901 & 237.2 & 8.2 & 40694.8 & 39668 & 494.6 & 8.5 & 40699.7 & 39660 & 534.9 & 8.8\\
			kro124p & 36230 & \textbf{39213.7} & 37538 & 947.5 & 15.4 & 41637.5 & 39438 & 1094.7 & 15.8 & 41608.3 & 39400 & 1116.8 & 15.7 \\			
			rbg323 & 1326 & 1640.9 & 1615 & 30.4 & 243.6 & \textbf{1634.7} & 1599 & 34.6 & 245.1 & 1639.7 & 1600 & 31.1 & 247.0\\
			\hline
		\end{tabular}
	}
	\caption{Results of the proposed IBA, DFA and DICA for the TSP and ATSP.}
	\label{tab:ResultsTSP3}
\end{table}

The conclusions that can be obtained from this second table are similar than the ones drawn previously. In this case, the IBA is the technique with the better performance, obtaining better results in the 67.56\% of the instances (25 out of 37), being worse in 8 cases. Overall, IBA has outperformed DFA and DICA in the 72.72\% of the TSP cases, and in the 60\% of the ATSP instances, getting worse results in 4 cases of each problem. Technique by technique, the IBA has performed better than the DFA in the 70.27\% of the cases (26 out of 37). Additionally, the IBA has outperformed the DICA in the 83.78\% of the instances (31 out of 37). Besides this, as in the comparison with the other techniques, the IBA has reached the optimal solution in more occasions that the DFA and DICA, outperforming these techniques also in this respect. Finally, the reflections about the robustness and the runtimes are the same as in the previous analysis.

\subsection{Statistical analysis and convergence behaviour analysis}
\label{sec:est}

Two different statistical tests have been conducted with the results obtained in previous subsections in order to obtain rigorous and fair conclusions. The guidelines given by Derrac et al. in (\cite{derrac2011practical}) have been followed to perform this statistical analysis. First of all, the Friedman's non-parametric test for multiple comparisons has been used to check if there are any significant differences among all the techniques. For the TSP problem (Table \ref{tab:results_friedmanTSPATSP}), the resulting Friedman statistic has been 65.525. Taking into account that the confidence interval has been stated at the 99\% confidence level, the critical point in a $\chi^2$ distribution with 5 degrees of freedom is 15.086. Since 65.525$>$15.086, it can be concluded that there are significant differences among the results reported by the five compared algorithms, being IBA the one with the lowest rank. Finally, regarding this Friedman's test, the computed p-value has been 0.0. On the other hand, the resulting Friedman statistic for the ATSP has been 27.761. Because 27.761$>$15.086, it can be concluded also for the ATSP that there are significant differences, being the IBA the best technique. In this case, the computed p-value has been 0.000041.

To evaluate the statistical significance of the better performance of IBA, the Holm's post-hoc test has been conducted using IBA as control algorithm. The unadjusted and adjusted p-values obtained through the application of Holm's post-hoc procedure can be seen in Table \ref{tab:results_holmsTSPATSP}. Analyzing this data, and taking into account that all the p-values are lower than 0.05, it can be concluded that IBA is significantly better for the TSP at a 95\% confidence level. On the other hand, for the ATSP, the IBA is significantly better than the GA, IDGA, ESA and DICA, and better, but not significantly, than the DFA.

\begin{table}[htb]
	\centering
	\scalebox{0.7}{
		\begin{tabular}{|c|c||c|c|}\hline
		\multicolumn{2}{|c||}{TSP} & \multicolumn{2}{c|}{ATSP} \\
		\hline
			Algorithm&Ranking & Algorithm&Ranking\\\hline
				IBA&1.4545 & IBA&1.8333\\
				ESA&3.5455 & ESA&3.5333\\
				GA&5.5682 & GA&4.9667\\
				IDGA&4.5455 & IDGA&4.5\\
				DFA&2.5909 & DFA&2.7667\\
				DICA&3.2955 & DICA&3.4\\
			\hline
		\end{tabular}
	}
	\caption{Average rankings returned by the Friedman's non-parametric test for the TSP and ATSP problems.}
	\label{tab:results_friedmanTSPATSP}
\end{table}

%

\begin{table}[htb]
	\centering
	\scalebox{0.7}{
		\begin{tabular}{|c|c c||c|c c|}\hline
		\multicolumn{3}{|c||}{TSP} & \multicolumn{3}{c|}{ATSP} \\
		\hline
		Algorithm & Unadjusted $p$& Adjusted $p$ & Algorithm & Unadjusted $p$& Adjusted $p$\\
		\hline
		GA&0&0 & GA&0.000005&0.000023\\
		IDGA&0&0 & IDGA&0.000095&0000379\\
		ESA&0.00021&0.00063 & ESA&0.012827&0.0.03848\\
		DICA&0.0011&0.0022 & DICA&0.021827&0.043654\\
		DFA&0.043951&0.043951 & DFA&0.171857&0.171857\\
		\hline
		\end{tabular}
	}
	\caption{Unadjusted and adjusted p-values obtained for the TSP and ATSP through the application of Holm's post-hoc procedure using IBA as control algorithm.}
	\label{tab:results_holmsTSPATSP}
\end{table}


To conclude the whole analysis of the results, and with the aim of making a deeper analysis, the convergence behavior shown by the IBA is compared next with the ones shown by the ESA and the DFA. We have selected the ESA and the DFA for this comparison because they are the most similar techniques in terms of average results quality with respect to IBA. In Table \ref{tab:convergenceTSP}, the average number of objective function evaluations needed to reach the final solution for each instance is shown (in thousands), as well as the standard deviations.

\begin{table}[tb]
	\centering
	\setlength{\tabcolsep}{2pt}
	\scalebox{0.7}{
		\begin{tabular}{| l || r r | r r | r r |}
		  \hline
		  \multicolumn{1}{|c||}{Instance} & \multicolumn{2}{c|}{IBA} & \multicolumn{2}{c|}{ESA} & \multicolumn{2}{c|}{DFA}\\
			\hline
			\hline Name & Avg. & S. dev. & Avg. & S. dev. & Avg. & S. dev.\\
			\hline
			Oliver30 & \textbf{2.17} & 0.48 & 23.91 & 3.49 & 3.38 & 1.45\\
			Eilon50 & 22.8 & 8.42 & 94.37 & 32.31 & \textbf{17.66} & 8.68\\
			Eil51 & \textbf{15.37} & 14.13 & 85.91 & 22.60 & 17.56 & 6.93\\
			Berlin52 & \textbf{20.07} & 6.02 & 128.26 & 49.82 & 23.68 & 7.31\\
			St70 & 72.67 & 30.38 & 216.08 & 77.49 & \textbf{69.56} & 32.82\\
			Eilon75 & \textbf{116.56} & 65.21 & 273.23 & 89.97 & 173.50 & 83.06\\
			Eil76 & \textbf{91.53} & 30.89 & 262.89 & 81.37 & 164.18 & 69.60\\
			KroA100 & \textbf{739.86} & 177.58 & 784.84 & 192.21 & 812.56 & 126.93\\
			KroB100 & \textbf{461.05} & 159.51 & 729.83 & 260.82 & 813.68 & 127.31\\
			KroC100 & 872.51 & 199.52 & \textbf{726.35} & 228.34 & 835.79 & 145.82\\
			KroD100 & \textbf{600.31} & 220.75 & 689.49 & 230.11 & 875.74 & 234.70\\
			KroE100 & \textbf{602.94} & 103.73 & 791.76 & 219.53 & 843.72 & 197.93\\
			Eil101  & \textbf{512.73} & 122.32 & 598.11 & 120.32 & 617.83 & 148.16\\
			Pr107 & 679.07 & 118.14 & \textbf{661.97} & 135.71 & 713.90 & 206.01\\
			Pr124 & 1602.51 & 436.32 & \textbf{1446.91} & 345.61 & 1589.71 & 399.07\\
			Pr136 & 2866.60 & 927.02 & \textbf{2318.20} & 655.24 & 2763.80 & 883.35\\
			Pr144 & 4361.11 & 1421.23 & \textbf{3678.46} & 943.12 & 4097.09 & 1349.64\\
			Pr152 & 4853.19 & 1639.27 & \textbf{3853.91} & 1037.53 & 4769.37 & 1709.60\\
			Pr264 & 6375.46 & 1864.76 & \textbf{6096.45} & 1749.12 & 6686.39 & 2009.73\\
			Pr299 & \textbf{6597.94} & 2001.91 & 6731.23 & 2067.71 & 7016.91 & 2364.28\\
			
			Pr439 & 8346.85 & 2739.64 & \textbf{8006.91}& 2996.35 & 8736.28 & 3069.18\\
			Pr1002 & 12103.73 & 4964.3 & \textbf{11038.32} & 4722.71 & 12843.60 & 5207.21\\
			
			\hline
			\hline
			br17 & 0.31 & 0.04 & 5.65 & 0.63 & \textbf{0.29} & 0.03\\
			ftv33 & 14.66 & 9.43 & 52.85 & 11.75 & \textbf{13.98} & 7.81\\
			ftv35 & \textbf{12.78} & 4.42 & 50.80 & 14.90 & 14.12 & 5.16 \\
			ftv38 & 25.53 & 9.68 & 49.89 & 13.30 & \textbf{23.43} & 10.51 \\
			p43 & \textbf{11.87} & 4.15 & 50.81 & 9.12 & 13.80 & 5.05\\
			ftv44 & \textbf{41.84} & 21.27 & 81.54 & 38.03 & 45.04 & 20.65\\
			ftv47 & 47.81 & 24.32 & 83.71 & 32.03 &\textbf{46.65} & 26.86\\
			ry48p & \textbf{42.11} & 13.32 & 79.45 & 19.52 & 49.16 & 18.70\\
			ft53 & \textbf{61.68} & 30.66 & 99.30 & 39.52 & 65.25 & 29.23\\
			ftv55 & \textbf{71.81} & 28.74 & 124.85 & 53.85 & 75.19 & 32.29 \\
			ftv64 & \textbf{89.86} & 40.15 & 143.89 & 65.42 & 93.40 & 42.95 \\
			ftv70 & \textbf{134.98} & 50.30 & 170.91 & 73.23 & 146.76 & 57.61 \\
			ft70 & \textbf{131.59} & 56.91 & 180.94 & 77.21 & 136.95 & 63.29\\
			kro124p & \textbf{438.59} & 101.62 & 641.49 & 163.99 & 451.02 & 134.79\\			
			rbg323 & \textbf{6948.06} & 1393.96 & 7316.76 & 2031.78 & 7006.54 & 1681.71\\
			\hline
		\end{tabular}
	}
	\caption{Convergence of IBA, ESA and DFA for TSP and ATSP, expressed in thousand of objetive function evaluations}
	\label{tab:convergenceTSP}
\end{table}

Analyzing the results shown in Table \ref{tab:convergenceTSP}, we can conclude that IBA outperforms both ESA and DFA also in terms of convergence. Overall, the IBA has shown a better performance in the 62.85\%of the instances (22 out of 35). Specifically, it can be concluded that the IBA performs better than the ESA in instances with, approximately, 100 nodes or less. This behavior has not been shown in the case of DFA, where the IBA has proved to be better in general terms. This fact provides an advantage to the IBA, since it can obtain better results needing less number of objective function evaluations.

As a final conclusion, we can say that using similar functions and parameters, the proposed IBA outperforms the other alternatives needing similar and acceptable runtimes and showing a better robustness and convergence. In addition, the improvements shown are significant in most cases. For this reason, we can say that the presented IBA is a promising approximation method to solve the TSP and ATSP, meeting, in this respect, the main objective of this study.

\section{Conclusions and Further Work}
\label{sec:conc}

In this work we have presented the first Discrete Bat Algorithm for solving the Traveling Salesman Problem and the Asymmetric Traveling Salesman Problem. In addition, we have proposed an improved version of the basic BA. In this Improved Bat Algorithm, bats are endowed with some kind of ``intelligence". This intelligence makes the bats follow different patterns of movement depending on the point of the solution space in which they are located. In order to prove that the proposed IBA is a promising approximation method to solve the TSP and ATSP, we have compared its performance along 37 instances with the one of two basic BAs. Furthermore, the results obtained by the IBA have been compared with those obtained by five different metaheuristics: a genetic algorithm, an evolutionary simulated annealing an island based distributed genetic algorithm, a discrete firefly algorithm and a discrete imperialist competitive algorithm. Additionally, three statistical tests have been conducted along the paper with the obtained outcomes: the Student's $t$-test, Holm's test and the Friedman test. Overall, the IBA has demonstrated a great performance for the TSP and ATSP, outperforming all the other alternatives, being the improvements significant in most of the cases.

Both TSP and ATSP problems are standard discrete problems; however, the conclusions obtained in this study cannot be generalized to other discrete problems. For this reason, as a future work, we intend to develop some additional versions of the proposed IBA to solve other routing problems. Our planned work includes the application of the IBA for the Capacitated Vehicle Routing Problem (\cite{ralphs2003capacitated}), and more complex routing problems, such as Rich Vehicle Routing Problems (\cite{caceres2014rich}). Additionally, we are aware of the large number of existing meta-heuristics in the literature. The comparison of the IBA with the five selected techniques is enough to prove that the proposed metaheuristic is a promising one. Nevertheless, we think that a wider experimentation with additional techniques can be valuable for the scientific community. In addition, we are planning to compare the proposed technique with exact methods and commercial solvers, using metrics such as the computational time or the number of explored solutions. These techniques are not similar to the IBA in terms of concepts and philosophy. However, we think such comparison will be very useful and can provide some insight into these methods.

\section*{Acknowledgement}
This project was supported by the European Union’s Horizon 2020 research and innovation programme   through the TIMON: Enhanced real time services for optimized multimodal mobility relying on cooperative networks and open data project (636220). As well as by the projects TEC2013-45585-C2-2-R from the Spanish Ministry of Economy and Competitiveness, and PC2013-71A from the Basque Government.

\bibliographystyle{plain}

\end{document}